\documentclass[11pt]{article}
\usepackage{coling2020}
\usepackage{times}
\usepackage{url}
\usepackage{latexsym}
\usepackage{graphicx}
\usepackage{multirow}
\usepackage{amsmath}
\usepackage{amssymb}
\usepackage{tabularx}
\usepackage{caption}
\usepackage{subcaption}

\colingfinalcopy 

\newcommand{\citep}[1]{\cite{#1}}
\newcommand{\citet}[1]{\newcite{#1}}
\newcommand{\citeauthor}[1]{\cite{#1}}

\title{Fine-tuning BERT for Low-Resource Natural Language Understanding via Active Learning}

\author{
    Daniel Grie\ss haber \\ Institute for Applied Artificial Intelligence (IAAI) \\ Hochschule der Medien Stuttgart \\ Nobelstra\ss e 10 \\ 70569 Stuttgart \\
    \texttt{griesshaber@hdm-stuttgart.de}
    \And Johannes Maucher \\ IAAI \\ Hochschule der Medien Stuttgart \\ Nobelstra\ss e 10 \\ 70569 Stuttgart \\ \texttt{maucher@hdm-stuttgart.de}
    \AND Ngoc Thang Vu \\ Institute for Natural Language Processing (IMS) \\ University of Stuttgart \\ Pfaffenwaldring 5B \\ 70569 Stuttgart \\
    \texttt{thangvu@ims.uni-stuttgart.de}
}

\date{}

\begin{document}
\maketitle
\begin{abstract}
    Recently, leveraging pre-trained Transformer based language models in down stream, task specific models has advanced state of the art results in natural language understanding tasks. However, only a little research has explored the suitability of this approach in low resource settings with less than 1,000 training data points. In this work, we explore fine-tuning methods of BERT - a pre-trained Transformer based language model - by utilizing pool-based active learning to speed up training while keeping the cost of labeling new data constant. Our experimental results on the GLUE data set show an advantage in model performance by maximizing the approximate knowledge gain of the model when querying from the pool of unlabeled data. Finally, we demonstrate and analyze the benefits of freezing layers of the language model during fine-tuning to reduce the number of trainable parameters, making it more suitable for low-resource settings.
\end{abstract}

\section{Introduction}
\blfootnote{
    %
    %

    %
    %
     \hspace{-0.65cm}  
     This work is licensed under a Creative Commons
     Attribution 4.0 International Licence.
     Licence details:
     \url{http://creativecommons.org/licenses/by/4.0/}.
    %
    %
}
Pre-trained language models have received great interest in the natural language processing (NLP) community in the last recent years \citep{dai_semi-supervised_2015,radford_improving_2018,howard_universal_2018,baevski_cloze-driven_2019,dong_unified_2019}.
These models are trained in a semi-supervised fashion to learn a general language model, for example, by predicting the next word of a sentence \citep{radford_improving_2018}.
Then, transfer learning \citep{pan_survey_2010,ruder-etal-2019-transfer,moeed-etal-2020-evaluation}
can be used to leverage the learned knowledge for a down-stream task, such as text-classification \citep{NIPS2005_2843,aggarwal_survey_2012,Reimers:2019:ACL,10.1007/978-3-030-32381-3_16}.

\citet{devlin_bert:_2019} introduced the  ``Bidirectional Encoder Representations from Transformers'' (BERT), a pre-trained language model based on the Transformer architecture \citep{vaswani_attention_2017}. BERT is a deeply bidirectional model that was pre-trained using a huge amount of text with a masked language model objective where the goal is to predict randomly masked words from their context \citep{taylor_cloze_1953}. The fact is, BERT has achieved state of the art results on the ``General Language Understanding Evaluation'' (GLUE) benchmark \citep{wang_glue:_2018} by only training a single, task-specific layer at the output and fine-tuning the base model for each task.
Furthermore, BERT demonstrated its applicability to many other natural language tasks since then including but not limited to sentiment analysis \citep{sun-etal-2019-utilizing,xu-etal-2019-bert,li-etal-2019-exploiting}, relation extraction \citep{baldini-soares-etal-2019-matching,10.1007/978-3-030-32236-6_65} and word sense disambiguation \citep{huang-etal-2019-glossbert,hadiwinoto-etal-2019-improved,huang-etal-2019-glossbert}, as well as its adaptability to languages other than English \citep{martin-etal-2020-camembert,antoun-etal-2020-arabert,agerri-etal-2020-give}.
However, the fine-tuning data set often contains thousands of labeled data points.
This plethora of training data is often not available in real world scenarios \citep{tan_empirical_2008,wan_using_2008,salameh_sentiment_2015,fang_model_2017}.

In this paper, we focus on the low-resource setting with less than 1,000 training data points.
Our research attempts to answer the question if pool-based active learning can be used to increase the performance of a text classifier based on a Transformer architecture such as BERT.
That leads to the next question: How can layer freezing techniques \citep{yosinski_how_2014,howard_universal_2018,peters_tune_2019}, i.e. reducing the parameter space, impact model training convergence with fewer data points?

To answer these questions, we explore the use of recently introduced Bayesian approximations of model uncertainty \cite{gal_dropout_2015} for data selection that potentially leads to faster convergence during fine-tuning by only introducing new data points that maximize the knowledge gain of the model. To the best of our knowledge, the work presented in this paper is the first demonstration of combining modern transfer learning using pre-trained Transformer-based language model such as the BERT model with active learning to improve performance in low-resource scenarios.
Furthermore, we explore the effect of trainable parameters reduction on model performance and training stability by analyzing the layer-wise change of model parameters to reason about the selection of layers excluded from training.

The main findings of our work are summarized as follows: a) we found that the model's classification uncertainty on unseen data can be approximated by using Bayesian approximations and therefore, used to efficiently select data for manual labeling in an active learning setting; b) by analyzing layer-wise change of model parameters, we found that the active learning strategy specifically selects data points that train the first and thus more general natural language understanding layers of the BERT model rather than the later and thus more task-specific layers.

\section{Methods}

\subsection{Base Model}
In \citep{devlin_bert:_2019} a simple classification architecture subsequent to the output of the Transformer is used to calculate the cross-entropy of the classifier for a text classification task with $C$ classes.
Specifically, first, a dropout operation \citep{srivastava_dropout:_2014} is applied to the Transformer's last layer hidden state of the special \texttt{[CLS]} token that is inserted at the beginning of each document. The regularized output is then fed into a single fully-connected layer with $C$ output neurons and a softmax activation function to scale the logits of its output to probabilities of class association.

\begin{figure}[ht]
      \centering
      \includegraphics[width=\textwidth]{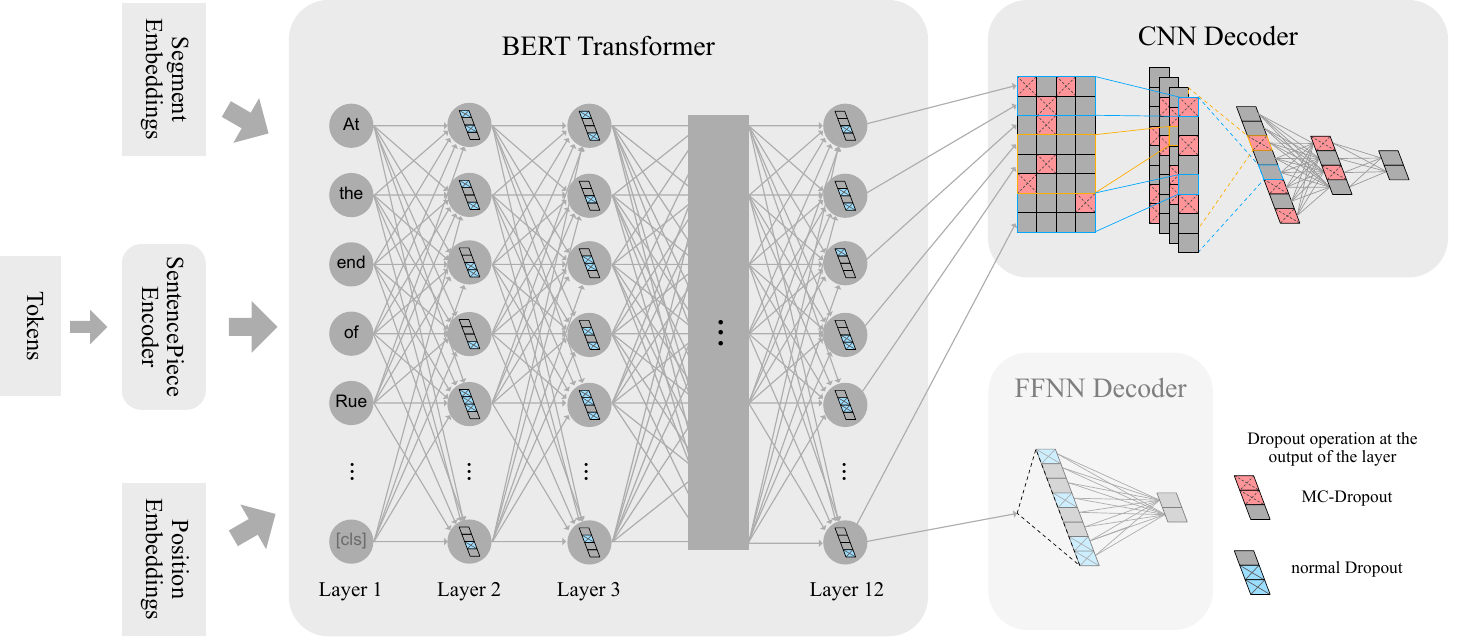}
      \caption{High-level overview of the architecture used in our experiments including a depiction of the simple FFNN decoder used by \citet{devlin_bert:_2019} highlighting where our model differs from the original experiment setup.}
      \label{fig:arch}
\end{figure}

In contrast, we use a more complex classification architecture based on a convolutional neural network (CNN) following \citet{kim_convolutional_2014}\footnote{We found that using a CNN is slightly better than using a simple feed forward neural network in our experiments.}. 
All hidden states in the last layer of the BERT model are arranged in a 2-dimensional matrix.
Then, convolutional filters of height ($3, 4, 5$) and length corresponding to the hidden state size ($786$) are shifted over the input to calculate $64$ 1-dimensional feature maps per filter size. These feature maps are then batch-normalized \citep{ioffe_batch_2015}, dropout regularized and global max-pooled before they are concatenated and fed into 2 fully-connected layers, each of which applies another dropout operation on its input \footnote{All implementation details can be found in the source code provided in
\ifcolingfinal
\raggedright\url{https://gitlab.mi.hdm-stuttgart.de/griesshaber/bayesian-active-learning}
\else
\raggedright\url{https://anonym}
\fi}.

\subsection{How to Select Data}
When labeled training data is sparse, but unlabeled in-domain data is readily available, manual labeling of all data is often not feasible due to cost.
In this scenario, it is advantageous to let the current model select $Q$ data points that it is most confused about from the pool of potential but yet unlabeled training elements ($U$) to be labeled by an human annotator. Then, they can be included in the training set $T_{new} = T_{old}\cup U_{{x\in U}_{arg max}(a(x, \mathcal{M}))\left<1,...,Q\right> }$ where $a(x, \mathcal{M})$ is a function predicting the knowledge gain of the model $\mathcal{M}$ when trained on the datum $x$.
Thus, the model is specifically trained on data that it can not yet confidently classify to maximize the knowledge gain in each training step, while keeping the cost of labeling new data constant.

We propose to estimate model prediction uncertainty by using Bayesian approximations as presented in \citet{gal_dropout_2015} for data selection process. The main idea is to leverage stochastic regularization layers (e.g. Dropout or Gaussian noises) that can be used to approximate model uncertainty ($\phi(x)$) on any datum ($x$) by performing multiple stochastic forward passes for each element in $U$.
This is implemented by applying the layer operation not only during training but also during the interference stage of the model.
Multiple forward passes of the model $\mathcal{M}$ with the same parameters ($\theta$) and inputs ($x$) thus yield different model outputs ($y$), as each pass samples a concrete model from a approximate distribution $q^*_\theta(\omega)$ that minimizes the Kullback-Leibler divergence to the true model posterior $p(\omega | T)$.
\citet{gal_dropout_2015} call this Monte-Carlo-Dropout (MC-Dropout) as the repeated, non-deterministic forward passes of the model can be interpreted as a Monte-Carlo process.

To decide which elements in $U$ are chosen, different acquisition functions ($a(x, \mathcal{M})$) could be used. In this work, we focus on the ``Bayesian Active Learning by Disagreement (BALD)'' \citep{houlsby_bayesian_2011} acquisition strategy because it demonstrated very good performance in comparison to other strategies in the experiments by \citet{gal_deep_2017} as well as in our own preliminary experiments.
BALD calculates the information gain for the model's parameters that can be achieved with the new data points, that is, the mutual information between predictions and parameters $\mathbb{I}\left[ y, \omega | x, T\right]$.
Hence, this acquisition function has maximum values for inputs that produce disagreeing predictions with high uncertainty.
This is equivalent to high entropy in the logits (the unscaled output of the network before the final softmax normalization) in a classifier model, as multiple (stochastic) forward passes of the model with the same input yield different classification results.
We use the same approximation of BALD that \citet{gal_deep_2017} used in their work (equation \ref{eqn:baldapprox}) where $\widehat{{p}}^{s}_{c}$ is the probability of class association (i.e. the softmax scaled logits) for input $x$ and class $c$ for one of $S$ samples $\widehat{\omega}_s \sim q^*_\theta(\omega)$ from the model's approximated posterior distribution, i.e. $\widehat{{p}}^{s} = \mathrm{softmax}(\mathcal{M}(y | x, \theta, \widehat{\omega}_s))$.

\begin{equation}
\label{eqn:baldapprox}
\begin{split}
&a_{BALD}(x, \mathcal{M}) = \\*
&-\sum_{c\in C}{\left(\frac{1}{S} \sum_{s=1}^{S}{ \widehat{{p}}^{s}_{c} }\right)\;
                log\left(\frac{1}{S} \sum_{s=1}^{S}{ \widehat{{p}}^{s}_{c} }\right)}
+ \frac{1}{S} \sum_{c\in C,s=1}^{s=S}{ \widehat{{p}}^{s}_{c}\;
                                    log\,\widehat{{p}}^{s}_{c}  }
\approx \mathbb{I}\left[ y, \omega | x, T\right]
\end{split}
\end{equation}

As a baseline, we compare the BALD strategy with random sampling of elements in $U$. That is, $a_{Rand}(x, \mathcal{M}) = unif[0,1)$, where $unif[a,b)$ is a sample from the uniform distribution in the half-closed interval $[a,b)$.

When using the BALD acquisition function, we sample for $S=50$ forward passes and add $Q=100$ data points with the highest model uncertainty according to the calculated BALD scores. When randomly acquiring new data points for the baseline, no forward passes are needed ($S=0$) while the number of acquisitions $Q$ stays constant. The MC-Dropout layers
that apply the dropout operation also during a stochastic forward pass of the model, are only placed in the classification architecture, thus the base model is used unaltered with regular dropout layers only active during the training phase. 
In our CNN architecture, the dropout is only applied in the penultimate layer of the network. We use the same dropout rate of $0.1$ as \citet{devlin_bert:_2019} in the decoder. This means that the stochastic forward passes only affect the output of the layers after the first MC-Dropout layer in the model. Thus, only a single and relativley expensive pass through the transformer is needed in each learning step while the required multiple passes though the subsequent classifiers are comparatively cheap.  

\subsection{How to Fine-tune Models}
\noindent \textbf{Reducing the Number of Trainable Parameters}
Freezing of parameters can be useful when fine-tuning a complex model, as it effectively lowers the number of parameters that need to be tuned during training \citep{howard_universal_2018}.
This is especially relevant in the low-resource setting with a small amount of training data, since the pre-trained BERT\textsubscript{BASE} model with $\sim\!110M$ parameters is over-parametrized for such small data sets.
However, since freezing parameters also lowers the adaptability of a model \citep{peters_tune_2019}, it is crucial to determine which parameters are frozen and which can be fine-tuned during training to not negatively affect the model's performance.

To our best knowledge no prior work has considered the training set size as a dependent parameter.
This parameter is especially important in low-resource settings.
Therefore, we will conduct experiments to visualize the model's performance in dependence of training data size with different sets of layers that are frozen during training.
We denote the number of frozen layers with $F$. For positive values, the layers in the half-closed integer interval $\left[ 0\;..\;F\right)$ are frozen, while negative values represent the interval $\left[ 12\!+\!F\;..\;12\right)$, i.e. the last $-F$ layers of the BERT model.
Layers of the classification architecture are always trainable during training because they are initialized randomly and thus need to be tuned.
We also visualize the change of model parameters in each layer during fine-tuning to reason about the choice of frozen layers.
Results of these experiments are presented in section \ref{sec:freezeresults}.

\noindent \textbf{Analyzing Changes in Layer-Parameters}
To analyze the changes in model parameters during the fine-tuning of the BERT model, we capture the initial state of all model parameters before starting the training.
This includes all pre-trained model weights, as well as the randomly initialized parameters of the CNN denoted by $\theta_0$.
We can then snapshot the same parameters after training for 3 epochs denoted by $\theta_3$.
The change in the model parameters can thus be described by $\Delta\theta = \theta_3 - \theta_0$.
Since the number of parameters is very high ($7,087,872$ for every layer of the BERT model while the exact number of parameters of the CNN depends on the output configuration of the model), we aggregate the change of parameters per layer by calculating the mean absolute difference (MAD) of the parameters.
The mean absolute difference in layer $x$ can be described by equation \ref{eqn:mad} where ${{\theta_e}_x}_n$ is the value of the $n$-th parameter in layer $x$ after training epoch $e$.
\begin{equation}
\label{eqn:mad}
\begin{split}
\overline{\left|\Delta\theta_x\right|}
&= \frac{1}{N}\sum_{i=0}^{N}{\left|\Delta{{\theta}_x}_n\right|}  \\*
&= \frac{1}{N}\sum_{i=0}^{N}{\left|{{\theta_0}_x}_n-{{\theta_3}_x}_n\right|}
\end{split}
\end{equation}

\section{Experiment Setup}
\noindent \textbf{Base Model}
The BERT model, as introduced by \citet{devlin_bert:_2019}, is used to create contextualized embeddings in all experiments.
Specifically, the pre-trained language model \textbf{BERT\textsubscript{BASE}}\footnote{Available under \raggedright\url{https://storage.googleapis.com/bert_models/2018_10_18/uncased_L-12_H-768_A-12.zip}} with 12 layers, 768 hidden states and 12 self-attention heads made available online by the authors is used.

\noindent \textbf{Data Sets}
Similar to the experiments presented by \citet{devlin_bert:_2019}, we use the ``GLUE Multi-Task Benchmark'' \citep{wang_glue:_2018} consisting of multiple NLP classification tasks including sentiment analysis and textual entailment to evaluate the performance of the different model configurations. Contrary to \citeauthor{devlin_bert:_2019}, we report the average accuracy of $N=3$ runs instead of the maximum achieved value for each setting. By doing so, we are able to analyze training stability by comparing the distribution of the results over different runs.



\noindent \textbf{Low-Resource Scenarios}
To simulate the low-resource setting in the GLUE tasks, we repeatedly evaluate the model's performance after training for $3$ epochs on a subset of the available data.
Since data points in the training set provided by the GLUE Benchmark are already shuffled, the subset $S_x$ simply contains the first $x$ data points ($S_x=\{s_1, s_2, \dots, s_x\}$) to ensure the same data selection between experiments.

\section{Results}
\label{sec:results}
\begin{figure}[ht]
\begin{minipage}{.45\textwidth}

    \centering
    \includegraphics[width=\columnwidth]{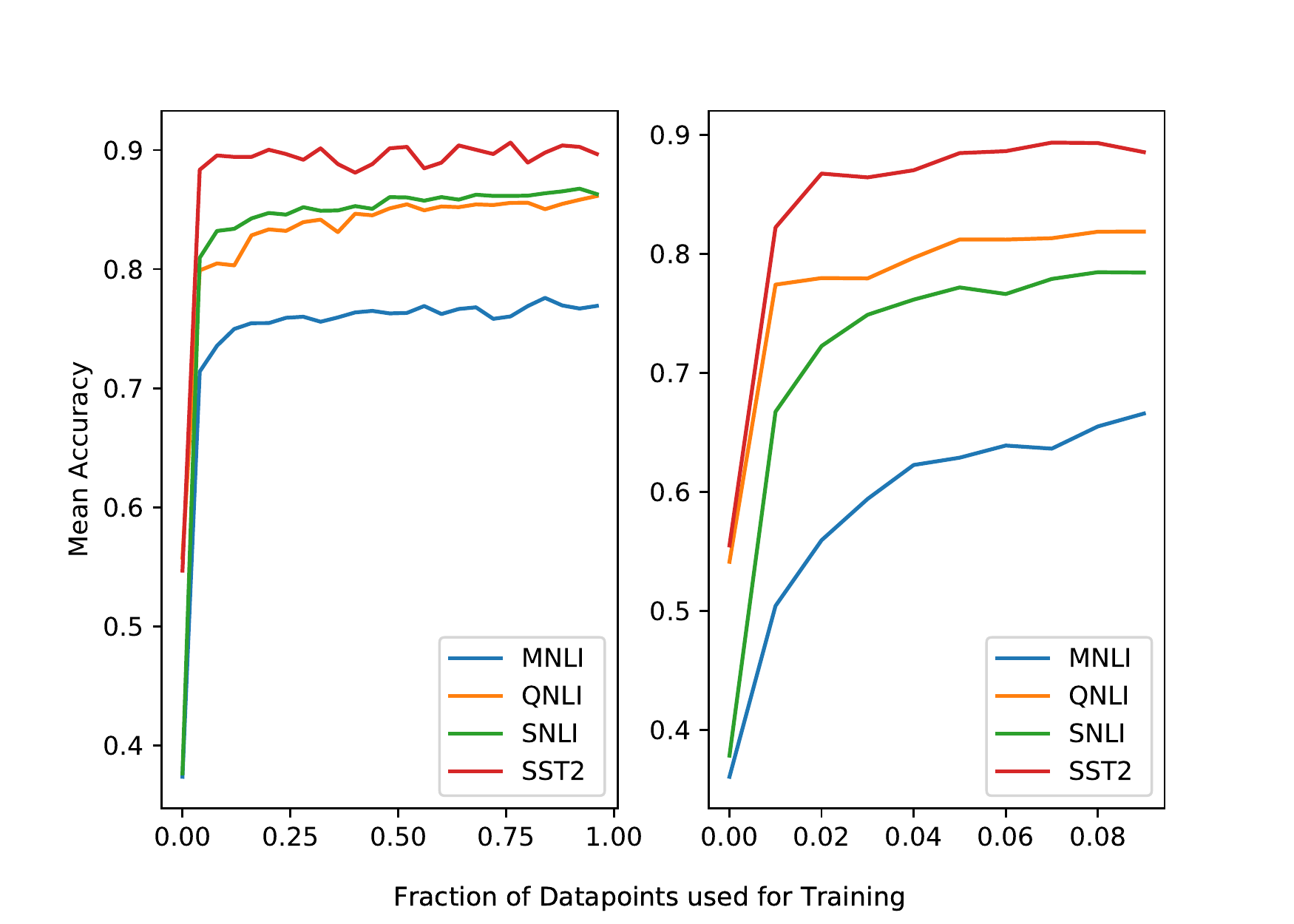}
    \caption{Accuracies of the BERT model when adapted on $100\%$ and $10\%$ (using random aquisition) of the available training data respectively. }
    \label{fig:res-size}

\end{minipage}
\hspace{0.09\textwidth}
\begin{minipage}{.45\textwidth}

    \centering
    \includegraphics[width=\columnwidth]{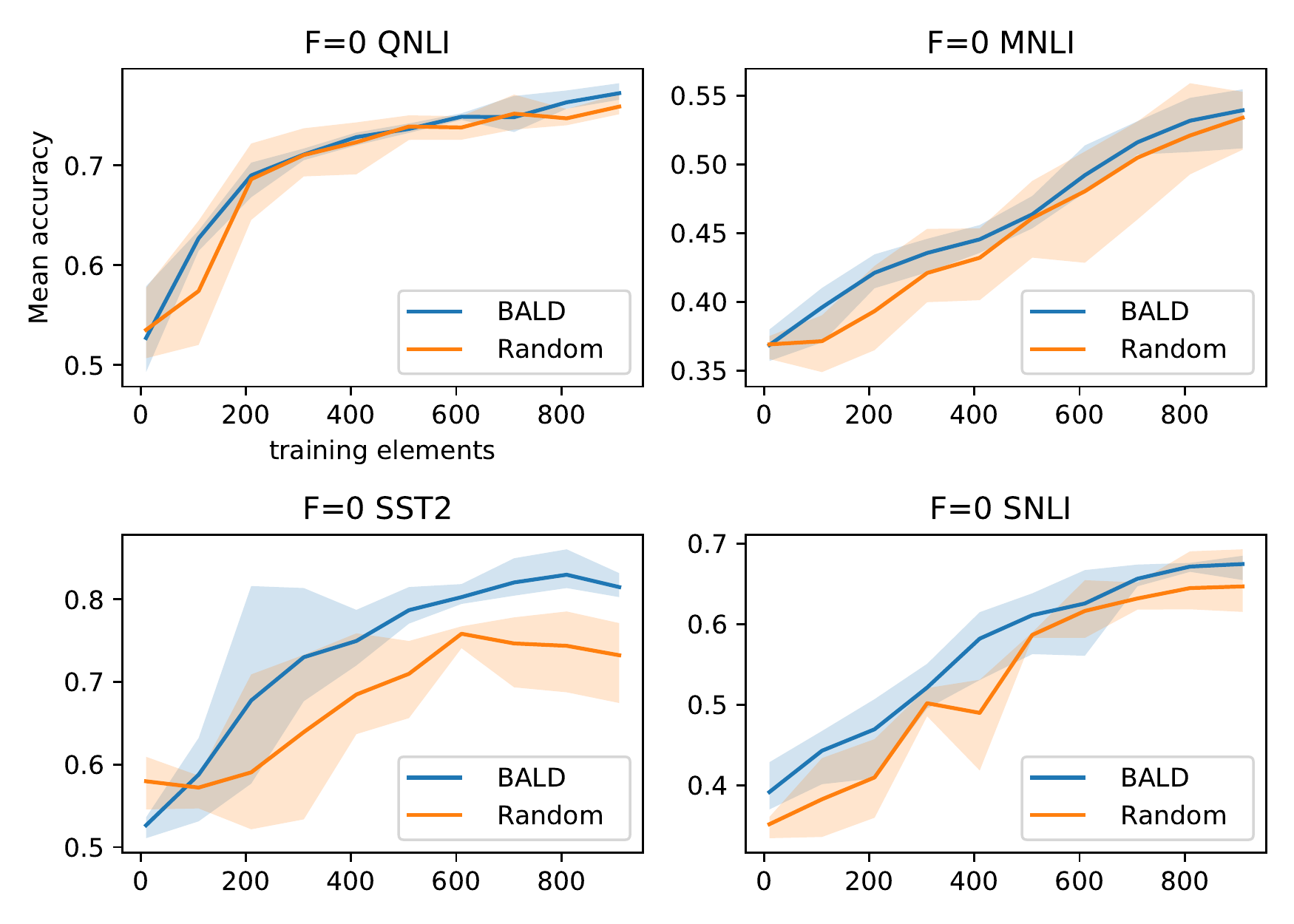}
    \caption{Model accuracies when adapted with data selected with the BALD acquisition strategy vs. random data selection.}
    \label{fig:res-activelearn}

\end{minipage}
\end{figure}

\subsection{Effect of Training Data Size}

As expected, all models benefit from an increased number of training elements until reaching comparable performance to the results reported by \citeauthor{devlin_bert:_2019} when trained on the whole data set.
However, Figure \ref{fig:res-size} shows that, while the accuracy is generally increasing for all experiments when introducing more training data, the model has a much larger increase in performance at the beginning, while adding another partition of the data set to an already large training set can only marginally improve the model's accuracy.
Adding more data, however, generally reduces the variance in model performance between runs when randomly initializing the classification architecture's parameters.
Due to these observations and for comparability of the results between data sets, the experiments in the following subsections are all performed on an initial training set of $S_{10}$ and reevaluated repeatably for $9$ iterations after adding another partition of $100$ elements until the final subset $S_{910}$ is reached. While at this point the performance has not yet converged to the final accuracies for any dataset, the performance changes drastically in these early training stages, which is ideal for our experimental comparison of training speed increases.

\subsection{Active Learning}
The active learning setting requires a pool of unlabeled data to pick the next training elements to acquire. Ideally, the knowledge gain $a(x, \mathcal{M})$ of all elements in the dataset would be approximated, however due to the need of multiple forward passes for each element in the pool to approximate $a$ only a subset $U = S_{20,000} \setminus S_{10} $ was chosen for our experiments to seed up training.
Thus, the size of $U$ can be seen as a hyperparameter to speed up training with the trade off of the risk to exclude highly relevant data points from the available pool data.
The size of $U$ was chosen by increasing it for multiple runs until the performance increase leveled off and acquisition times were still acceptable.
Since training with the random strategy does not have the same trade-off, all available data points in the pool can be used for the acquisition without lengthening the training.

The results in Figure \ref{fig:res-activelearn} show a better mean accuracy for all models and generally across all training set sizes when the BALD strategy is used, compared to picking random training elements from $U$.
Another noteworthy observation is the lower spread of model accuracy between experiments when training data is selected based on the active learning strategy.
These results show a strong indication that the Monte-Carlo approximation of model uncertainty works for state of the art Transformer architectures like the BERT model and can improve training performance when an active learning scenario in a low-resource setting is feasible.

\newcolumntype{Y}{>{\centering\arraybackslash}X}

%

\subsection{Layer Freezing to Reduce Number of Parameters}
\label{sec:freezeresults}

\begin{figure}[ht]

    \begin{subfigure}{\textwidth}
      \centering
      \includegraphics[width=\columnwidth]{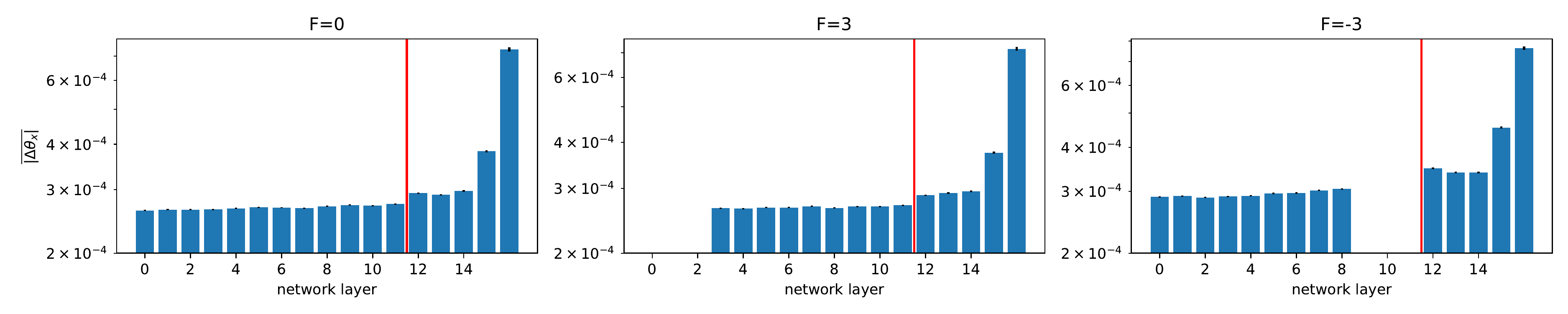}
      \caption{random sampling of new training elements from $U$}
      \label{fig:sfig1}
    \end{subfigure}
    \newline%
    \begin{subfigure}{\textwidth}
      \centering
      \includegraphics[width=\columnwidth]{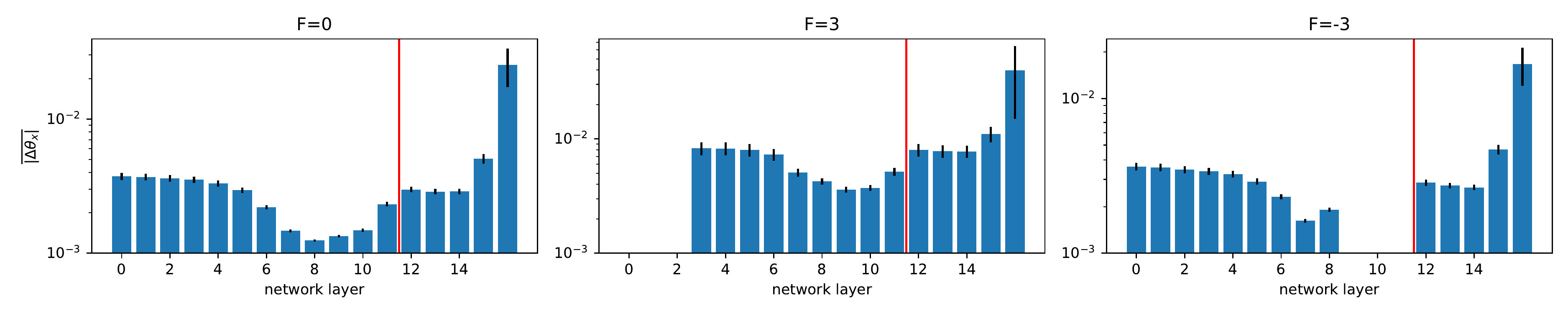}
      \caption{selective acquisition of elements with maximum BALD score from $U$}
      \label{fig:sfig2}
    \end{subfigure}

    \caption{Visualization of the mean absolute difference of parameters when training a classifier model on the QNLI data set. The blue bars indicate the variance over all weight changes. The red line indicates the border between the BERT model and the decoder CNN. Layers 12, 13 and 14 are the parallel convolutional layers, 15 and 16 the fully-connected hidden and output layers.}
    \label{fig:res-weightdelta}

\end{figure}

Figure \ref{fig:res-weightdelta} visualizes the mean absolute difference of model parameters on a per-layer basis when fine-tuning on the QNLI data set.
Note that the frozen layers have a mean absolute difference of $0$ as the training does not alter any parameters in these layers.
When randomly sampling from $U$ is used to choose new data elements, layers closer to the output of the model generally have a higher change in their parameters during training which is in line with the findings by  \citet{yosinski_how_2014}, \citet{howard_universal_2018} and \citet{hao_visualizing_2019} which show that, while fine-tuning a Transformer-based language model, the front layers have a more general language understanding while the later layers capture more task specific concepts and thus need to be trained more. 
However, if the active learning strategy is used, layers closer to the output of the BERT model have a smaller value change while the relative mean absolute difference in the layers of the CNN is comparable between the two strategies.
This indicates that the active learning strategy specifically selects data points that train the first and thus more general layers of the BERT model. 

Freezing of layers with a smaller overall change in parameters during training may be beneficial in low-resource scenarios, since this reduces the number of parameters in the model equally while potentially enabling more freedom in the layers that need to be tuned more.
Based on this hypothesis, we compare the model's performance when freezing different numbers of layers starting on both, the input and output of the BERT model.

\begin{table}[hb]

\begin{minipage}{.45\textwidth}
    \newcommand{\specialcell}[2][c]{%
    \begin{tabular}[#1]{@{}c@{}}#2\end{tabular}
    }

\begin{tabularx}{\columnwidth}{|c|Y|Y|Y|Y|}
	\hline
    $F$ &                MNLI                &                QNLI                &                SST2                &                SNLI                 \\
	\hline
	 0  &  $0.53 \pm 0.021$  &  $0.76 \pm 0.010$  &  $0.78 \pm 0.059$  &  $0.67 \pm 0.015$  \\ \hline
	 3  &  $0.51 \pm 0.021$  &  $0.78 \pm 0.003$  &  $0.80 \pm 0.045$  &  $0.69 \pm 0.002$  \\ \hline
	-3  &  $0.52 \pm 0.010$  &  $0.78 \pm 0.002$  &  $0.84 \pm 0.013$  &  $0.69 \pm 0.008$  \\ \hline
	 6  &  $0.51 \pm 0.024$  &  $0.75 \pm 0.014$  &  $0.81 \pm 0.006$  &  $0.63 \pm 0.014$  \\ \hline
	-6  &  $0.47 \pm 0.020$  &  $0.77 \pm 0.010$  &  $0.64 \pm 0.094$  &  $0.64 \pm 0.067$  \\ \hline
\end{tabularx}
    \caption{Mean accuracies of the model after training on $910$ data points using the BALD acquisition strategy.}
    \label{tab:res-freezeaccuracies}
    \end{minipage}
    \hspace{0.09\textwidth}
\begin{minipage}{.45\textwidth}

  \centering
    \begin{tabularx}{\columnwidth}{|c|Y|Y|Y|Y|}
	\hline
    $F$ &  MNLI   &  QNLI   &  SST2   &  SNLI    \\
	\hline
	 0  & $0.054$ & $0.049$ & $0.108$ & $0.061$ \\
	 3  & $0.043$ & $0.032$ & $0.078$ & $0.032$ \\
	-3  & $0.038$ & $0.024$ & $0.047$ & $0.028$ \\
	 6  & $0.050$ & $0.033$ & $0.071$ & $0.082$ \\
	-6  & $0.050$ & $0.050$ & $0.113$ & $0.161$ \\ \hline
\end{tabularx}
    \caption{Mean width of the confidence intervals, i.e. the difference between upper and lower limit, over different runs using the BALD acquisition strategy.}
    \label{tab:res-decoder}

\end{minipage}
\end{table}
Table \ref{tab:res-freezeaccuracies} shows a general increase in model performance when freezing 25\% of the BERT's layers, indicating that a reduction of parameters in the low-resource setting is indeed beneficial.
However, with 50\% of the layers fixed during training, the average performance of the model decreases again, in many cases even below the baseline where all parameters are trainable.
This is an indicator that the BERT model needs fine-tuning and freezing a large ratio of layers may result in a model that can not adapt adequately to the task and is in line with the results of \citet{peters_tune_2019}.

\begin{figure}[ht]
    \centering
    \includegraphics[width=\columnwidth]{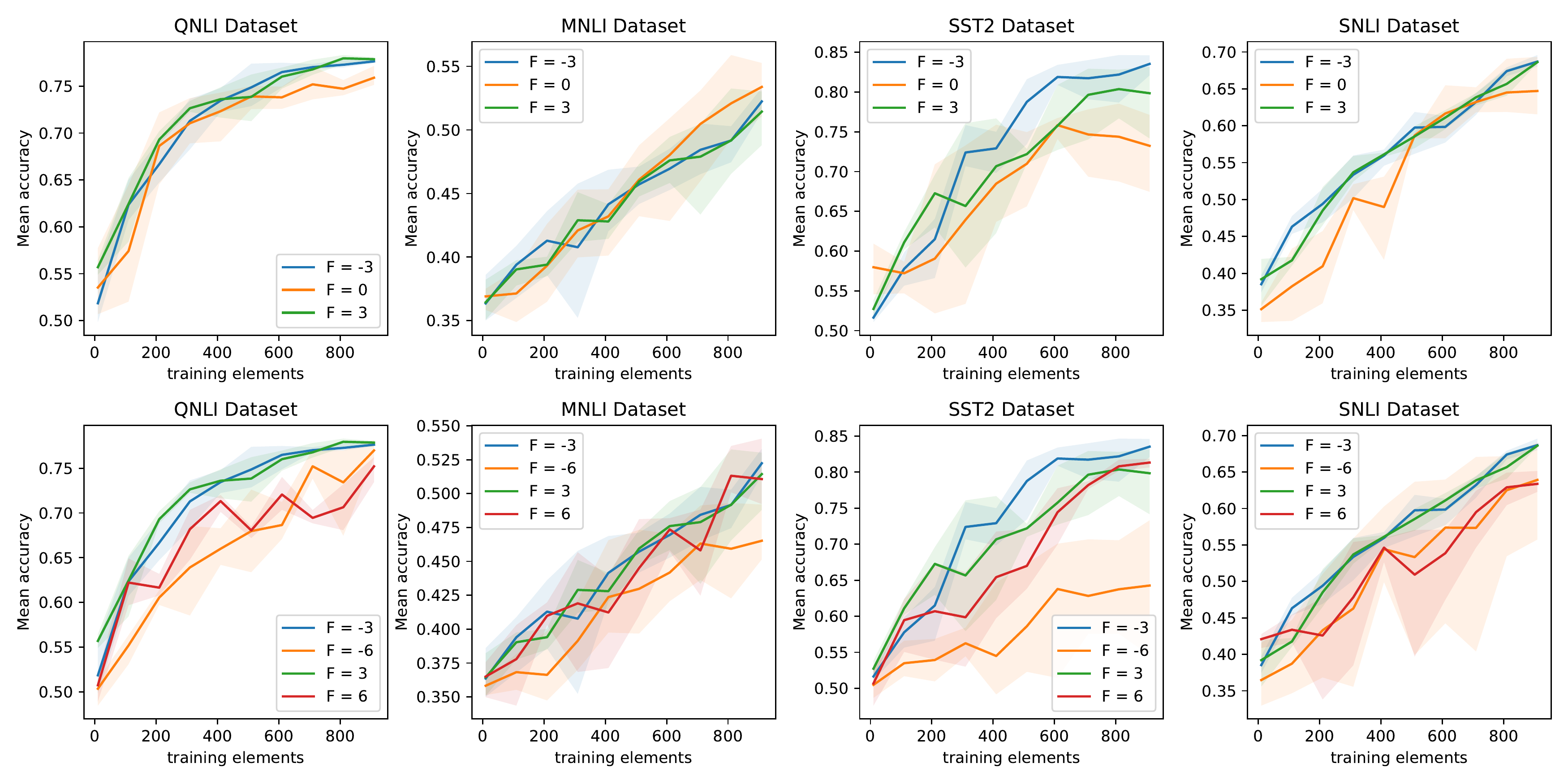}
    \caption{The effect of layer-freezing on the stability of the training during active learning using the BALD strategy. The semi-transparent area in the chart is the confidence interval over multiple runs of the configuration, while the solid line denotes the average accuracy.}
    \label{fig:res-freeze}
\end{figure}

Figure \ref{fig:res-freeze} and Table \ref{tab:res-decoder} also show the mean accuracy and bounds of these experiments over a range of different numbers of training elements.
The average accuracy of the models after multiple training trials do not show any conclusive advantage of freezing the later BERT layers over the the ones in the front.
However, for all data sets the models are more stable during the training over different runs, indicated by a lower average width of the bounds when using $F=-3$.

\section{Qualitative Analysis}
\noindent \textbf{Overall Observation}
Figure \ref{fig:res-alaqu} visualizes the number of training elements in $T$ grouped by their label ${}_c$ after each acquisition iteration of picking $100$ elements from $U$.
As the labels in the data sets and thus the subset $U$ are equally distributed, we expect the same equal distribution when sampling randomly from the pool elements, which is apparent in the lower row of Figure \ref{fig:res-alaqu} where the training data was actually randomly chosen from the pool of data.
In contrast, we observed that when using the BALD acquisition with non-deterministic forward passes, the distribution shows a stronger bias to a particular class.
This bias increases when sampling more pool data, whereas the difference between the biggest and smallest class in $T$ stays constant during random acquisition (see table \ref{tab:res-acqdiff}).
Furthermore, when randomly sampling, the class with the most or least training elements (${\mathrm{arg\,min}}_{c}\left|T_c\right|$, ${\mathrm{arg\,max}}_{c}\left|T_c\right|$) is changing in many iterations, whereas in the case where active learning is used, this is constant for the most part and changing, if any, only in the first iterations.

\noindent \textbf{Which Class?}
As the goal of active learning is to maximize the knowledge gain of the model with minimum cost, the choice of data the acquisition strategy selects from $U$ may give further insight into the models understanding of the input data.
One characteristic when applying active learning on the MNLI data set, that is apparent in Figure \ref{fig:res-alaqu}, is that samples from the \texttt\small\texttt{neutral} class are queried the most. It might indicate that the model is most confused about the samples from this class. This observation may be justified by the fact that the hypothesis stated in that datum may be unrelated to the associated premise, confusing the model.

Another observation is that the model queries data points from the \texttt\small\texttt{contradiction} class more often than data points from the \texttt\small\texttt{entailment} class at the beginning of the training, while at the end number of queries from the different classes is almost equal.
This indicates that, at the start of training, the model is more confused about contradictions, which may be explained by the fact that a contradiction can differ from an entailment only by a single negation at the correct place in the input, making it harder to differentiate from an entailment.
An example of negated statements in a contradiction can be seen in the last row of Table \ref{tbl:queryexamples} which also shows that the model indeed gave this example a high BALD score.

Table \ref{tbl:queryexamples} shows some of the examples of the MNLI dataset with their calculated BALD scores in the middle of the training after 5 acquisition iterations.
These examples show another noteworthy behaviour of the active learner.
Samples from the entailment class where many words in the two input sentences match or and have a similar wording in general, get a low BALD score, indicating that the model is already confident that it is able to correctly classify those examples.
The same is true for input pairs that differ in wording and contradict themselves.
However, entailing pairs where the wording is mostly different between the two inputs, as well as contradicting pairs with similar phrasing get a much higher BALD score and thus are more likely to be sampled during the acquisition phase.
The model at this point in training thus seems to already have learned to perform its task confidently on the \textit{simpler} examples and can thus concentrate more on the \textit{non-trivial} data-points.


\begin{table}[h!t]
\begin{minipage}{.45\textwidth}
    \begin{tabularx}{\columnwidth}{cc|c|c|c|c|}
    \cline{3-6}
    \multicolumn{1}{l}{}                                & \multicolumn{1}{l|}{} & \multicolumn{4}{c|}{Data set}                            \\ \cline{3-6}
    \multicolumn{1}{l}{}                                &                       & \multicolumn{2}{c|}{BALD} & \multicolumn{2}{c|}{Random} \\ \cline{3-6}
                                                        &                       & MNLI        & QNLI        & MNLI          & QNLI        \\ \hline
    \multicolumn{1}{|c|}{\multirow{9}{*}{\rotatebox{90}{$\left|T\right|$}}} & 110                     & 13          & 30          & 3             & 4           \\ \cline{2-6}
    \multicolumn{1}{|c|}{}                              & 210                     & 19          & 24          & 3             & 2           \\ \cline{2-6}
    \multicolumn{1}{|c|}{}                              & 310                     & 20          & 16          & 1             & 2           \\ \cline{2-6}
    \multicolumn{1}{|c|}{}                              & 410                     & 29          & 18          & 2             & 2           \\ \cline{2-6}
    \multicolumn{1}{|c|}{}                              & 510                     & 30          & 36          & 9             & 4           \\ \cline{2-6}
    \multicolumn{1}{|c|}{}                              & 610                     & 28          & 40          & 6             & 6           \\ \cline{2-6}
    \multicolumn{1}{|c|}{}                              & 710                     & 37          & 44          & 3             & 2           \\ \cline{2-6}
    \multicolumn{1}{|c|}{}                              & 810                     & 34          & 44          & 4             & 4           \\ \cline{2-6}
    \multicolumn{1}{|c|}{}                              & 910                     & 25          & 36          & 9             & 4           \\ \hline
    \end{tabularx}
    \caption{Differences in number of elements in the largest and smallest group in the trainset: $\Delta\left|T\right| = \mathrm{max}(\left|T_c\right|) - \mathrm{min}(\left|T_c\right|)$}
    \label{tab:res-acqdiff}
\end{minipage}
\hspace{.09\columnwidth}
\begin{minipage}{.45\textwidth}

\centering
\includegraphics[width=\columnwidth]{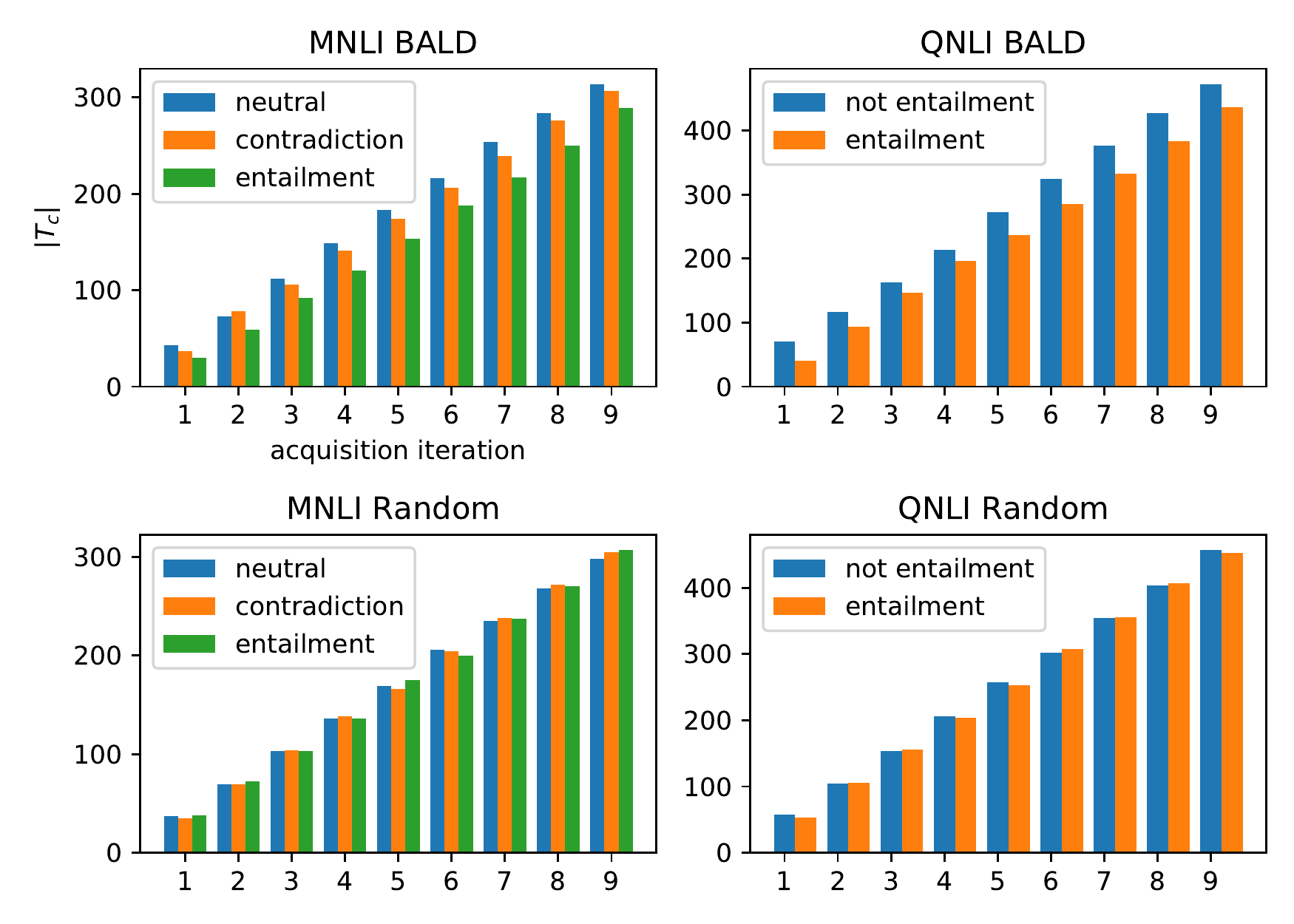}
\captionof{figure}{Distribution of training elements in $T$ for different acquisition strategies with the BERT model. $\left|T_c\right|$ is the number of elements in $T$ labeled with ${}_c$}
\label{fig:res-alaqu}
\end{minipage}
\end{table}

\begin{table}[h!t]
    \begin{tabularx}{\textwidth}{|c|r|X|X|l|}
    \hline
queried & $a_{BALD}(x, \mathcal{M})$ & $x_1$                                                                                                                                                                       & $x_2$                                                    & $y$           \\ \hline
no      & $-8.2\times 10^8$          & The view of the illuminated mont at night is spectacular from the other side of the bay.                                                                                    & The view of the lit up mont at night is wonderful.       & entailment    \\ \hline
no      & $-5.9\times 10^8$          & At the end of Rue des Francs-Bourgeois is what many consider to be the city's most handsome residential square, the Place des Vosges, with its stone and red brick facades. & Place des Vosges is constructed entirely of gray marble. & contradiction \\ \hline
yes     & $6.2\times 10^8$           & The grounds, like those of any other private house after nightfall, seemed untenanted.                                                                                      & The grounds were empty.                                  & entailment    \\ \hline
yes     & $8.6\times 10^8$           & And Forrest saying: "We don't give medals, Sergeant.                                                                                                                        & Forrest said they gave medals ever week.                 & contradiction \\ \hline
yes     & $5.2\times 10^8$           & I did not wonder at John objecting to his beard.                                                                                                                            & John had no objections to his beard.                     & contradiction \\ \hline
\end{tabularx}
    \caption{Examples from the MNLI Dataset with the two input sentences and the associated class label. The first column shows whether or not this data point was queried by the active learner during the complete training, $a_{BALD}(x, \mathcal{M})$ is the calculated BALD score of the input after 5 acquisition iterations ($\left|T\right|=510$).}
    \label{tbl:queryexamples}
\end{table}

\section{Related Work}
\label{sec:related}
\noindent \textbf{Low-resource NLP} Previous work in low-resource NLP tasks includes feature-engineering \citep{tan_empirical_2008} which requires a recurring effort when adapting to a new data set.
Another approach is to transfer knowledge across domains to increase the amount of data that is available for training \citep{zoph_transfer_2016,nguyen_transfer_2017,kocmi_trivial_2018,yang_transfer_2017,gupta_semi-supervised_2018}.
One of these approaches relied on adversarial training \citep{goodfellow_generative_2014} to learn a domain adaptive classifier \citep{ganin_domain-adversarial_2016} in another domain or language where training data was plentiful while ensuring that the model generalizes to the low-resource domain \citep{chen_adversarial_2017}.
However, these approaches have not used a pre-trained generic language model, but perform pre-training for each task individually.

\noindent \textbf{Adapting pre-trained models} The effectiveness of transfer learning in low-resource settings was previously demonstrated for machine translation \citep{zoph_transfer_2016,nguyen_transfer_2017,kocmi_trivial_2018}, sequence tagging \citep{yang_transfer_2017} and sentiment classification \citep{gupta_semi-supervised_2018}. However, all this prior work does not use general LMs but transfers knowledge from a pre-trained model to a close to target corpus.
Some previous work has analyzed the performance behavior of the BERT model in different scenarios.
\citet{hao_visualizing_2019} showed that a classifier that fine-tunes a pre-trained BERT model generally has wider optima on the training loss curves in comparison to models trained for the same task from scratch, indicating a more general classifier \citep{chaudhari_entropy-sgd:_2017,li_visualizing_2018,izmailov_averaging_2018}.
\citet{peters_tune_2019} examine the adaption phase of LM based classifiers by comparing fine-tuning and feature extraction where the LMs parameters are fixed.
In contrast, we focus on the low-resource setting where less than 1,000 data points are available for fine-tuning.

\noindent \textbf{Layer freezing in deep Transformers} Experiments by \citet{yosinski_how_2014} indicated that the first layers of a LM capture a more general language understanding, while later layers capture more task-specific knowledge.
With this motivation, \citet{howard_universal_2018} introduced \textit{gradual unfreezing} of the Transformer layers during each epoch, beginning with the last layer.
\citet{hao_visualizing_2019} analyzed the \textit{loss surfaces} in dependency of the model parameters before and after training and came to the same conclusion that lower layers contain more transferable features. However, none of the work has considered the training set size as a dependent parameter as our experiments presented in this paper.

\noindent \textbf{Active learning in NLP}
There is some prior work regarding active learning for NLP tasks using deep neural networks.
\citet{zhang_active_2017} explored pool-based active learning for text classification using a model similar to our setting. However, they used word-level embeddings \citep{mikolov_efficient_2013} and focus on representation learning, querying pool points that are expected to maximize the gradient changes in the embedding layer.
\citet{shen_deep_2017} used active learning for named entity recognition tasks. They proposed a acquisition strategy named \textit{Maximum Normalized Log-Probability} which is a normalized form of the \textit{Constrained Forward-Backward} confidence estimation \citep{culotta_confidence_2004,culotta_reducing_2005}.
Using this strategy, they achieved on-par performance in comparison to a model using the BALD acquisition function and MC Dropout without needing multiple forward passes.
However, this approach is not suitable for any arbitrary model architecture but requires conditional random fields (CRFs) for the approximation of model uncertainty.

\section{Conclusion}
In this paper, we evaluated the performance of a pre-trained Transformer model - BERT - in an active learning scenario for text classification in low-resource settings.
We showed that using Monte-Carlo Dropout in the classification architecture is an effective way to approximate model uncertainty on unlabeled training elements.
This technique enables us to select data for annotation that maximize the knowledge gain for the model fine-tuning process. Experimental results on GLUE data set show that it improves both model performance and training stability.
Finally, in order to improve the efficiency of the fine-tuning process with a small amount of data, we explored the reduction of trainable model parameters by freezing layers of the BERT model up to a certain level of depth. Comparing the exclusion of layers in the front or the back of the BERT model from training, we found it to be advantageous for training stability when freezing the layers closest to the output.


\ifcolingfinal
\section*{Acknowledgments}
This research and development project is funded within the ”Future of Work” Program by the German Federal Ministry of Education and Research (BMBF) and the European Social Fund in Germany. It is implemented by the Project Management Agency Karlsruhe (PTKA). The authors are responsible for the content of this publication.
\fi

\bibliographystyle{coling}
\bibliography{coling2020}
\end{document}